# Ear Recognition With Score-Level Fusion Based On CMC In Long-Wave Infrared Spectrum

Umit KACAR
The Institute of Science and Technology
Istanbul Technical University
Istanbul, Turkey
umitkacar@itu.edu.tr

Murvet KIRCI
Department of Electronic and Communication Engineering
Istanbul Technical University
Istanbul, Turkey
ucerm@itu.edu.tr

*Abstract*— **Only a few studies have been reported regarding human ear recognition in long-wave infrared (LWIR) band. Thus, we have created ear database based on LWIR. We have called that the database is "LWIR MIDAS" consisting of 2430 records of 81 subjects. Thermal band provides seamless operation both night and day, robust against spoofing with understanding ear liveness and invariant to illumination conditions for human ear recognition. We have proposed to use different algorithms to reveal the distinctive features. Then, we have reduced the number of dimensions using subspace methods. Finally, the dimension of data is reduced in accordance with the classifier methods. After this, the decision is determined by the best sores or combining some of the best scores with matching fusion. The results have showed that the fusion technique was successful. We have reached 97.71% for rank-1 with 567 test probes. Furthermore, we have defined the perfect rank which is rank number when recognition rate reaches 100% in cumulative matching curve (CMC). This evaluation is important for especially forensics, for example corpse identification, criminal investigation etc.**

*Keywords*— **long-wave infrared, ear recognition, score level fusion, perfect rank, ear thermogram, ear spoofing.**

## I. INTRODUCTION

Security measures have been increased recently due to the terror and the violence. Cards predefined generally are read previously given in the security zone control at the entrance of the building access. However, when these cards are copied or stolen, security breaches occur. Security system on based biometrics is trending in order to prevent the aforementioned problems.

Biometrics data needs to be read quickly and easily about staff entering biometrics-based security systems. Therefore, in comparison with other biometrics, the contactless ear and face biometrics systems are at the forefront, especially in the military applications. The contactless biometrics system can also be used law enforcement, border surveillance, force protection forensic etc.

Furthermore, the biometrics system must be robust against the spoofing. Thermal camera largely overcomes it. Thermal infrared straightforwardly identifies with the thermal radiation from object, which relies on upon emissivity of the materials and the temperature of the item. The passive infrared band is further divided into mid-wave infrared (MWIR (3-5μm)) and long-wave infrared (LWIR (8-12μm)). Moreover, face and ear liveness can easily be detected by the thermal camera [1] and thermal images are invariant to illumination conditions. Visible band is not available for night time environments.

In addition, when the contactless biometrics system enables remote recognition is inadequate, a ubiquitous biometrics system will provide a cue such as a person's location using Global Positioning System (GPS) for identification of unauthorized protesting [2]. Thus, the biometrics system should be evaluated the perfect rank score in cumulative matching curve (CMC) besides the best score, called rank-1. Investigation person can coincide the lowest recognized subject in the biometrics systems. Thus, we have defined the perfect rank whose details are described in Section 5.

Comparing to face and ear biometrics, it is clear to see that ear biometrics has a prevalence regarding uniform color distribution, spatial resolution, constant and clear background, luminance less sensitive, and being non-sensitive to the components such as mimics, gesture, and cosmetics etc.

## II. RELATED WORK

Ear recognition database have been freely, licensed or costly created in the visible band until today [3-5]. Although thermal face recognition is focused [6-12], there are few studies on ear recognition in thermal band. In the literature, we have found three studies, including two LWIR and one MWIR band. First study was used the commercial tools, including the biofilter matching process with 14 subjects, each taken seven right ear thermal image in the same session [13]. It declared that the results were not sufficiently distinguishable under the same conditions (equal error rate (EER) 20.7%). At the same time this study attempted to find the temperature zones between the ear and the hair. Burge and Burger had proposed to segment the certain temperature zones to distinguish from the ear to the hair [14]. However, our work and above-mentioned [13] showed that temperature zones of the ear and the hair were intersected. Thus, this segmentation will not be possible for everyone. Second work is used PCA methods (EER 14%) with 75 subjects, each taken 15 right ear thermal image in the different days [15].

Last study based on MWIR is used PCA, ICA, LDA, SIFT, uLBP and LTP method with total 45 subjects, including each 30 subjects are taken six right and six left ear thermal image for training, each 45 subjects are taken one right and one left for test gallery and one right and one left for test probe in the time lapse session. The result shows, respectively, LTP and uLBP to yield the best scores [16].

Last study MWIR based ear database is pilot study and not public. Others need to be developed. As a result, there is a research-gap on thermal ear recognition. Thus, we have created ear database based on LWIR at Istanbul Technical University (ITU). We have called that the database is "LWIR MIDAS". This database includes 81 subjects, each taken 15 left and 15 right ear thermal image under 0-60 degrees rotation, some occlusion by hair and earring in the same session. The properties of this database are described in the Section 4.

## III. EAR IN THERMAL IMAGERY

Nowadays, with the increase in resolution thermal camera and the decrease in price, it has proliferated research on thermal imaging. It is possible to buy high definition thermal camera or LWIR camera using with the mobile phone[1].

### A. Ear Thermogram

Thermogram[2] of a nearby up perspective of a man's ear indicating varieties in temperature shown in Fig.1. The temperature keeps running from red (hottest) through yellow, green, cyan and blue to mauve (coldest).

The area of ear canal or concha (yellow color in Fig.1.) is largely hotter than helix, lobule and antihelix (blue color in Fig.1.). At the same time, the area of ear canal or concha is as hot as face, but helix, lobule and antihelix is generally colder than face temperature. Especially, if the ambient environment is pretty cold, this difference is quite clear. This scenario can facilitate to detect ear in thermal band against the visible band. The external anatomy of the ear is also illustrated in Fig.1.

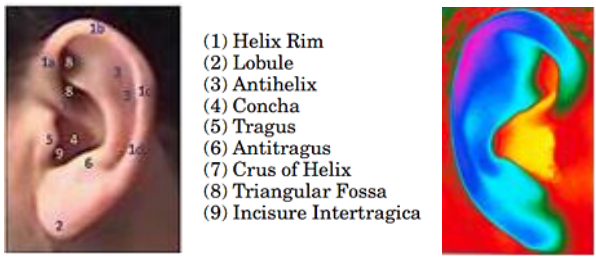

Fig. 1. Ear anatomy [3] (left), ear thermogram (right) [2]

### B. Ear Spoofing

The 2D ear biometrics systems can be deceived by a photograph or a video of the ear displayed on a screen of portable device. [17]. 3D ear recognition systems are also spoofed by 3D ear model or ear mask made of silicon, rubber etc. in Fig.2. Thermal camera largely overcomes spoofing. Face and ear liveness can easily be detected by the thermal camera. Owing to detecting heat, it seems to be difficult same thermal image which is perceived by thermal camera.

### C. Thermal images limitation

In spite of the fact that thermal images are valuable in face and ear recognition [6-16], there are a few restrictions for thermal band. There are different conditions under which the thermal characteristics for human can be changed.

Firstly, one of the changes that appear in the ear region is disease such as red ear syndrome etc. [18]. Secondly, the reason of change is human physiology. Blushing regularly happens when someone is ashamed or furious. Heart rate and breathing frequency increase during these circumstances, since more adrenaline is excreted by human body. The veins in face also open up to permit more blood to flow. Owing to flowing more blood than normal through the capillaries in face, face will seem redder, but less ear. The influence is significantly more prominent in people with white skin. Thirdly, peripheral conditions makes ear (especially helix, lobule, antihelix), hand, toe and nose which are flowing less blood, cold. Actually, the heating of the body is related to blood circulation. When the weather is cold, ears are colder than face shown as a Celsius. Other reasons are hot beverages, spicy nourishments, consumption of alcohol, heat and sun exposure and exercise in hot weather etc. [19].

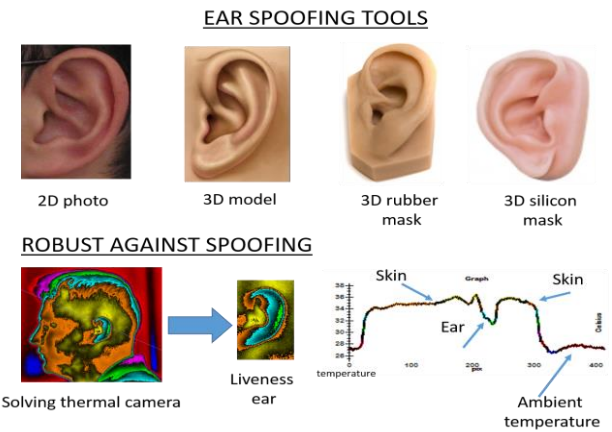

Fig. 2. Ear Spoofing

## IV. EXPERIMENTAL SETUP

Thermal videos are taken for every volunteer seated swivel chair between 0-180 degrees from left profile to right profile. Then, thermal video is converted the image between 60 degrees span for left and right ears. Owing to the more difficulty of recognition and identification in LWIR band, we did not focus ear detection. The problem of ear detection can be solved by manual cropping or automated methods with a cascaded AdaBoost based of Haar features [20]. Thus, left and right profile images are cropped only to remain ear region. The resolution of all the ear image is taken as 80x60.

---

[1] http://www.flir.com/flirone/content/?id=62910

[2] http://fineartamerica.com/featured/thermogram-of-a-close-up-of-a-human-ear-dr-arthur-tucker.html

## A. LWIR Camera

The thermal camera used in this study is a high resolution uncooled thermal GigE vision camera by Xenics in Fig.3. The high performance thermal imaging camera achieves frame rates up to 50 Hz at full 640 x 480 image resolution or higher in windowing mode. The detector includes a small 17μm pixel pitch and low 50 mK noise equivalent temperature difference (NETD) with germanium window. The spectral range of the camera is 8-14μm with 16 bit resolution. The camera was equipped with a 100 mm LWIR lens additionally provided by Xenics.

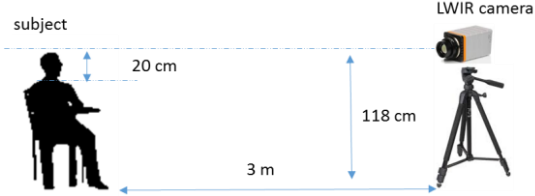

Fig. 3. Xenics Gobi 640 Gige uncooled LWIR camera.

## B. LWIR MIDAS ear database

LWIR MIDAS database includes left and right ear images of 81 individuals. There are 30 images from each individual, 15 from left ear and 15 from right ear. Images were taken indoors with 0 to 60 degrees rotation for left and 120 to 180 degrees for right ear. The distance between LWIR camera and subject is fixed to 3 meters. This ear database is called LWIR MIDAS ear database. The database was divided into three main sections: training set, testing galleries and testing probes, shown in Fig.4. Training set includes 50 subject with 7 samples image each individual. For testing, gallery is one image against 7 samples probe.

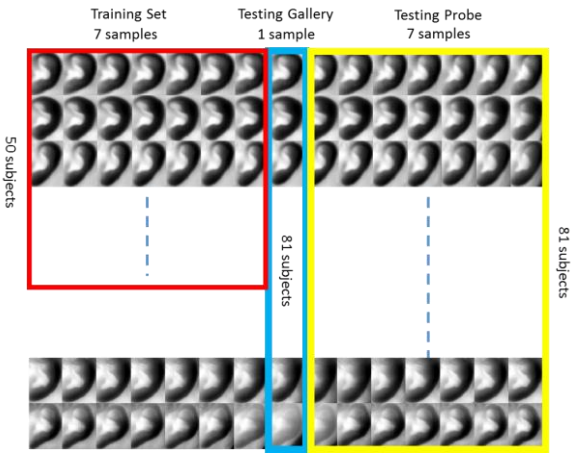

Fig. 4. Xenics Gobi 640 Gige uncooled LWIR camera.

## V. EXPERIMANTAL RESULTS

We analyze different ear identification algorithms proposed in the visible band and application of these methods in the LWIR band for left ears. These feature extraction processes are intensity-based such as, PCA [21], LDA [22], and DCVA [23]; texture-based such as uLBP [24], LPQ [25], and HOG [26]. Owing to extracting the discriminability of shape based information from low-quality 2D image, we did not use shape-based features [27]. There is no pre-processing on the images. It is used euclidean distance for PCA, LDA and DCVA based intensity and chi-square distance for uLBP, LPQ based texture.

TABLE I. EXPERIMENTAL RESULTS.

| Sequence Number | EAR BIOMETRICS METHODS | IDENTIFICATION RATE % | | | | | PERFECT RANK | VR EER (%) |
|---|---|---|---|---|---|---|---|---|
| | | RANK-1 (%) | RANK-2 (%) | RANK-3 (%) | RANK-4 (%) | RANK-5 (%) | | |
| 1 | PCA | 74.07 | 80.42 | 84.13 | 85.54 | 86.24 | 44 | 8.71 |
| 2 | LDA | 88.89 | 92.95 | 93.47 | 94.18 | 94.36 | 58 | 5.39 |
| 3 | DCVA | 93.12 | 96.30 | 97.00 | 98.06 | 98.41 | 31 | 2.92 |
| 4 | uLBP (8,2) | 87.65 | 90.12 | 91.53 | 92.77 | 93.30 | 35 | 4.50 |
| 5 | uLBP (16,2) | 88.36 | 91.53 | 93.12 | 93.47 | 93.47 | 33 | 3.91 |
| 6 | LPQ | 88.18 | 90.83 | 92.24 | 92.95 | 93.65 | 34 | 3.89 |
| 7 | HOG+LDA | 94.71 | 96.65 | 97.00 | 98.06 | 98.59 | 10 | 1.11 |
| 8 | HOG +DCVA | 92.42 | 95.24 | 97.53 | 99.47 | 99.65 | 7 | 1.71 |
| 9 | uLBP (8,2) +LDA | 93.65 | 95.77 | 97.35 | 98.24 | 98.41 | 13 | 1.24 |
| 10 | uLBP (16,2) +LDA | 89.95 | 94.36 | 96.30 | 97.53 | 97.88 | 19 | 1.81 |
| 11 | uLBP (8,2) +DCVA | 92.42 | 94.71 | 97.18 | 97.88 | 98.24 | 20 | 2.16 |
| 12 | uLBP (16,2) +DCVA | 91,36 | 94,53 | 96,30 | 97,36 | 97,88 | 32 | 2.30 |
| 13 | LPQ +LDA | 93.65 | 96.47 | 97.36 | 98.24 | 98.41 | 47 | 1.82 |
| 14 | LPQ +DCVA | 90.12 | 93.65 | 95.42 | 97.00 | 97.88 | 17 | 1.93 |

In this study, when the methods given in Table-1 are implemented, we have showed not only identification rate from rank-1 to rank-5 and equal error rate (EER) for the verification scenario, but also rank number for 100% of identification rate. We have also defined the perfect rank. It indicates the rank number when recognition rate reaches 100% in CMC curve. This evaluation is important for especially forensics, corpse identification, criminal investigation etc. Owing to this situation, all suspects up to the perfect rank of biometrics system should be investigated for identification.

The best results are HOG+LDA for rank-1, rank-2 and EER, but HOG+DCVA from rank-3 to rank-5. HOG+DCVA is the best method for minimizing the perfect rank number with seven against 81 class. Some of the best methods results are combined by matching score level fusion [28, 29]. Since matching scores are based on different scaling, the matching scores cannot be used or combined directly. Score normalization is required by changing over the scores into common similar scale or domain. Further, some of the best scores are normalized with min-max normalization technique between 0-1 and weighted, shown in Table-2. We have reached

97.71% for rank-1 and also increased from rank-2 to rank-5 with matching fusion methods. Feature extraction with different algorithms provides a variety perspectives of related data. The aim is to reveal the distinctive features. Finally, the decision is determined by the best sores or combining some of the best scores with matching fusion techniques. Computational load increases for matching fusion techniques, but the accuracy is higher.

TABLE II. MATCHING SCORE FUSION RESULTS.

| MATCHING FUSION | WEIGHTED | IDENTIFICATION RATE % ||||| PERFECT RANK |
|---|---|---|---|---|---|---|---|
| | | RANK-1 (%) | RANK-2 (%) | RANK-3 (%) | RANK-4 (%) | RANK-5 (%) | |
| 8+9 | .75+.25 | 96.12 | 98.77 | 99.47 | 99.65 | 99.82 | 6 |
| 8+9+13 | .63+.12+.25 | 97.71 | 99.30 | 99.82 | 99.82 | 99.82 | 8 |

## VI. CONCLUSION

Only a few studies have been reported regarding human ear recognition in LWIR band. The methods used in studies are simple and the performance is low. Owing to the research-gap in LWIR band, we have created ear database based on LWIR. We have called that the database "LWIR MIDAS". We proposed to use different algorithms to reveal the distinctive features. Then, we have reduced the number of dimensions using subspace methods. Finally, the dimension of data is reduced in accordance with the classifier methods. After this, the decision is determined by the best sores or combining some of the best scores with matching fusion. The results showed that the fusion technique was successful. We have reached 97.71% for rank-1 with 567 test probes. Furthermore, we have defined the perfect rank which is rank number when recognition rate reaches 100% in CMC curve. The person requested recognition may be in the perfect rank. Thus, the perfect rank should be minimized. In the future work, we will try to do fully automated methods for ear biometrics in LWIR band.